\RequirePackage{fix-cm}
\documentclass[smallextended]{svjour3}       
\smartqed  
\usepackage{graphicx}
\usepackage{amssymb}
\usepackage{amsmath}
\usepackage[linesnumbered,ruled,vlined]{algorithm2e}
\usepackage{color}

\begin{document}

\title{Nonlinear Regression Analysis Using Multi-Verse Optimizer
}


\author{Jayri Bagchi         \and
        Tapas Si 
}


\institute{J. Bagchi \at
              Department of Computer Science \& Engineering\\
              Bankura Unnayani Institue of Engineering\\
              Bankura--722146, West Bengal, India\\
              \email{jayribagchi@gmail.com}           
           \and
           T. Si \at
              Department of Computer Science \& Engineering\\
              Bankura Unnayani Institue of Engineering\\
              Bankura--722146, West Bengal, India\\
              \email{c2.tapas@gmail.com} 
}

\date{Received: date / Accepted: date}

\maketitle

\begin{abstract}
Regression analysis is an important machine learning task used for predictive analytic in business, sports analysis, etc. In regression analysis, optimization algorithms play a significant role in search the coefficients in the regression model. In this paper, nonlinear regression analysis using a recently developed meta-heuristic Multi-Verse Optimizer (MVO) is proposed. The proposed method is applied to 10 well-known benchmark nonlinear regression problems. A comparative study has been conducted with Particle Swarm Optimizer (PSO). The experimental results demonstrate that the proposed method statistically outperforms PSO algorithm.
\end{abstract}

\section{Introduction}
\label{intro}
Regression analysis is a statistical method to explain the relationship between independent and dependent variables or parameters and predict the coefficients of the function. Linear regression analysis involves those functions that are linear combination of the independent parameters whereas nonlinear regression analysis is a type of regression analysis in which given data is modeled with a function that is a nonlinear combination of multiple independent variables. Some examples on nonlinear regression functions are exponential, logarithmic, trigonometric, power functions.
Regression is the most important and widely used statistical technique with many applications in business and economics.
\\
\indent OZSOY et al.~\cite{Ref2} performed an estimation of nonlinear regression model parameters using PSO. This study compares the optimal and estimated parameters and its results hence show that the estimation of the coefficients using PSO yield reliable results. Mohanty~\cite{Ref9} applied PSO to astronomical data analysis. The results show that PSO requires tuning of few parameters compared to GA but is found to be slightly worse than GA. A case study  of PSO in regression analysis by Cheng et al.~\cite{Ref6} utilized PSO to solve a regression problem in the dielectric relaxation field. The results show that PSO with ring structure has a good mean solution than PSO with a star structure.
Erdogmus and Ekiz~\cite{Ref9} proposed a nonlinear regression analysis using PSO and GA for some test problems shows that GA shows better performance in estimating the values of the coefficients. Their work further shows that such heuristic optimization algorithms can be an alternative to classic optimization methods. Lu et al.~\cite{Ref3} performed a selection of most important descriptors to build QSAR models using modified PSO (PSO-MLR) and compared the results with GA (GA-MLR). The results reveal that PSO-MLR performed better than GA-MLR for the prediction set. Barmaplexis et al.~\cite{Ref4} applied multi-linear regression, PSO and artificial neural networks in the pre-formulation phase of mini-tablet preparation to establish an acceptable processing window and identify product design space. Their results show that DoE-MLR regression equations gave good fitting results for 5 out of 8 responses whereas GP gave the best results for the other 3 responses. PSO-ANNs was only to fit all selected responses simultaneously. Cerny et al.~\cite{Ref5} proposed a new type of genotype for Prefix Gene Expression Programming (PGEP). PGEP, improved from Gene Expression Programming (GEP) is used for Signomial Regression(SR). The method was called Differential Evolution-Prefix Gene Expression Programming (DE-PGEP) which allows for expression and constants to co-exist in the same vector spaced representation and be evolved simultaneously. Park et al.~\cite{Ref7} proposed PSO based Signomial Regression (PSR) to solve non-linear regression problems. Their work attempted to solve the signomial function by estimating the parameters using PSO. Mishra \cite{Ref12} evaluates the performance of Differential Evolution at nonlinear curve fitting. Results show that DE has been successful to obtain optimum results even if parameter domains were wide but it couldn't reach near-optimal results for the CPC-X problems which are the challenge problems for any nonlinear least-squares algorithm. Gilli et al.~\cite{Ref13} used DE, PSO and Threshold Accepting methods to estimate the parameters of linear regression. Yang et al.~\cite{Ref10} constructed the linear regression models for the symbolic interval-values data using PSO.
\\
\indent The objective of this paper is to perform a nonlinear regression analysis using the MVO algorithm~\cite{Ref1}.   The proposed method is applied to 10 well-known benchmark nonlinear regression problems. A comparative study is conducted with PSO~\cite{Ref8}. The experimental results with statistical analysis demonstrate that the proposed method outperforms PSO. 

\subsection{Organization of this paper}
The remaining of the paper is organized as follows: the proposed method is discussed in section~\ref{sec:1}. The experimental setup including the regression models and  dataset description is given in section~\ref{sec:2}. The results and discussion are given in section~\ref{sec:3}. Finally, the conclusion with future works is given in section~\ref{sec:4}.
\section{Materials \& Methods}
\label{sec:1}

\subsection{Regression Analysis}
Regression analysis is a statistical technique to estimate the relationships among the variables of a function. 
It is a commonly used method for obtaining the prediction function  for predicting the values of the response variable using predictor variables~\cite{Ref16}.
There are three types of variable in regression such as
\begin{itemize}
    \item The unknown coefficients or parameters, denoted as $\beta$, may be represent a scalar or a vector
    \item The independent variable or predictor variable, i.e., input vector $X=(x_1,x_2,\ldots,x_n)$
    \item The dependent variable or response variable, i.e., output $y$
\end{itemize}
The regression model in basic form can be defined as:
\begin{equation}
    y \approx f(x,\beta)
\end{equation}
where $\beta=(\beta_0, \beta_1, \beta_2,\ldots, \beta_m)$.\\
A linear regression model is a model of which output variable is the linear combination of coefficients and input variables and it is defined as~\cite{Ref15}:

\begin{equation}
    y=\beta_0+\beta_1x_1+\beta_2x_2+\cdots+\beta_nx_n+\xi
\end{equation}
where $\xi$ is a random variable, a disturbance that perturbs the output $y$.
A nonlinear regression model is a model of which output variable is the nonlinear combination of coefficients and input variables. The nonlinear regression model is defined as follows~\cite{Ref15}:

\begin{equation}
    y=f(x,\beta)+ \xi
\end{equation}
where $f$ is the nonlinear function.

In regression analysis, an optimizer is used to search the coefficients, i.e., parameters so that the model fits well the data. In the current work, the unknown parameters of different nonlinear regression models are searched using MVO algorithm. The MVO algorithm is discussed next.

\subsection{MVO Algorithm}
 Multi-Verse Optimizer is an optimization algorithm whose design is inspired by the multiverse theory in Physics~\cite{Ref1}. Multiverse theory in Physics states that there exist multiple universes and each universe possesses its own inflation rate which is responsible for the creation of stars, planets, asteroids, meteroids, black holes, white holes, wormholes, physical laws for that universe.


For a universe to be stable, it must have a minimum inflation rate. So the goal of the MVO algorithm is to find the best solution by reducing the inflation rate of the universes which is also the fitness value. Now, observations from multiverse theory show that universes with higher inflation rate have more white holes and universes with a low inflation rates have more black holes. So to have a stable situation, objects from white holes have to travel to black holes. Also the objects in each universe may travel randomly to the best universe through wormholes.


In MVO, each solution represents a universe and each variable to be an object in the universe. Further, the inflation rate is assigned to each universe which is proportional to the fitness value of each universe. MVO uses the concept of black holes and white holes for exploring search spaces and wormholes to exploit search spaces. When a tunnel is established between two universes, it is assumed that the universe with a higher inflation rate has more white holes and the universe with a lower inflation rate has more black holes. So universes exchange objects from white holes to black holes which improves the average inflation rates of all universes over the iterations. 
In order to mathematically model the above idea, the roulette Wheel mechanism is used that selects one of the universes with a high inflation rate to contain a white hole and allows objects from that universe to move into the universe containing a black hole and relatively low inflation rate. At every iteration, universes are sorted and one of them is selected by the roulette wheel to have a white hole.
Assuming that $U$ is the matrix of universes with $d$ parameters and $n$ candidate solutions. 

Roulette wheel selection mechanism based on the normalized inflation rate is illustrated as below:
\begin{equation} 
x_i(j)  =
\begin{cases}
 x_k(j) &  r_1<NI(U_i) \\
 x_i(j) & r_1>=NI(U_i)
\end{cases}
\end{equation}

Here $x_i(j)$ indicates $j$th parameter of $i$th universe, $U_i$ shows $i$th universe, $NI(U_i)$ is the normalized inflation rate of $i$th universe, $r_1$ is a random number in the interval $[0,1]$ and $x_k(j)$ is the $j$th parameter of the $k$th universe selected by roulette wheel mechanism.\\ 
As this is done with the sorted universes, so the universes with low inflation rates have a higher probability if sending objects through white/black holes.
Now in order to perform exploitation, it is considered each universe has wormholes to transport its objects randomly through space. In order to improve average inflation rates, it is assumed that wormhole tunnels are established between a universe and the best universe obtained so far. 
The formulation of the mechanism is:
\begin{equation}
x_i^j =
\begin{cases}
\begin{cases}
 x_j+TDR*((ub_j-lb_j)*r_4+lb_j) & r_3<0.5\\
 x_j-TDR*((ub_j-lb_j)*r_4+lb_j) & r_3>=0.5
\end{cases} & r_2<WEP\\
x_i^j & r_2>=WEP
\end{cases}
\end{equation}
Here $x_j$ indicates the $j$th parameter of best universe formed so far, $TDR$ and $WEP$ are coefficients, $lb_j$ is the lower bound of $j$th variable, $ub_j$ is the upper bound of $j$th variable, $x_i^j$ indicates the $j$th parameter of $i$th universe, and $r_2$, $r3$, $r_4$ are random numbers in $[0,1]$.

$WEP$ is the wormhole existence probability and $TDR$ is the traveling distance rate. $WEP$ is for defining the probability of the existence of wormhole. It is to be increased linearly over the iterations for better exploitation results. TDR defines the distance rate that an object can travel through the wormhole
to the best universe obtained so far. $TDR$ is decreased over the iterations to increase the accuracy of local search by the following rule:
\begin{equation}
    TDR=1-\frac{l^\frac{1}{p}}{L^\frac{1}{p}}
\end{equation}
where $l$ is the current iteration and $L$ is the maximum number of iterations. $p$ is the exploitation accuracy over iterations and generally, it is set to $6$.
The update rule of $WEP$ is as follows:
\begin{equation}
    WEP=WEP_{\min}+l*\left(\frac{WEP_{\max}-WEP_{\min}}{L}\right)
\end{equation}
where $WEP_{\min}$ and $WEP_{\max}$ indicate the minimum and maximum range of $WEP$.

\begin{figure}[!tbh]
\includegraphics[height=18cm, width=11cm]{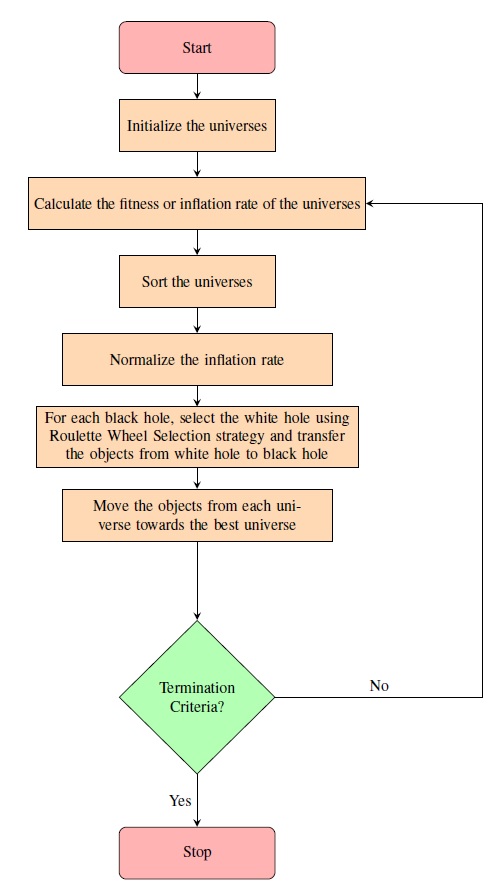}
\caption{Flowchart of MVO.}
\label{fig:mvo_flowchart}       
\end{figure}

\subsection{Regression Analysis Using MVO}
In this work, MVO is used to search the unknown parameters ($\beta$) of the nonlinear regression model. Let assume a model has $m$ number of parameters and then the dimension of the universe in MVO is $m$. The $i$th universe  is represented by $X_i=(x_1,x_2,\ldots,x_m)$. The inflation rate of the universe is the objective function value. The mean square error (MSE) is used as an objective function in this work. The MSE is calculated as follows:
\begin{equation}
   MSE=\frac{1}{N}\sum_{i=1}^{N}(y_{i}- y_{i}{'})^{2}
\end{equation}
where $y_i$ and $y_{i}{'}$ are the target and predicted output of the $i$th input data respectively. MVO algorithm minimizes MSE to fit the data. $N$ is the number of samples in the dataset.

\section{Experimental Setup}
\label{sec:2}
\subsection{Regression Model \& Dataset Description}
In this work, 10 regression model is analyzed and the datasets for the models have been collected from~\cite{Ref14}. 
The description of different regression models and their dataset is given in Table~\ref{tab:model}.\\

\begin{table}[!tbh]
\caption{Regression model and dataset description}
\label{tab:model}
\begin{tabular}{p{0.3cm}p{1.0cm}p{4.4cm}p{2.5cm}p{1.2cm}}
\hline\noalign{\smallskip}
Sl. No. & Name & Model & No. of coefficients & No. of samples\\
\noalign{\smallskip}\hline\noalign{\smallskip}
1 & Misra1a & $\beta_1(1-exp(-\beta_2x))$ & 2 & 14\\
2 & Gauss1 & $\beta_1exp(-\beta_2x)+\beta_3\frac{-(x-\beta_4)^2}{\beta_5^2}+\beta_6\frac{-(x-\beta_7)^2}{\beta_8^2}$ & 8 & 250\\
3 & DanWood & $\beta_1x^{\beta_2}$ & 2 & 6\\
4 & Nelson & $exp(\beta_1-\beta_2x_1exp(-\beta_3x_2))$ & 3 & 128\\
5 & Lanczos2 & $\beta_1exp(-\beta_2x)+\beta_3exp(-\beta_4x)+\beta_5exp(-\beta_6x)$ & 6 & 24\\
6 & Roszman1 & $\beta_1-\beta_2x-\frac{arctan\frac{\beta_3}{x-\beta4}}{\pi}$ & 4 & 25\\
7 & ENSO & $\beta_1+\beta_2cos\frac{2\pi x}{12}+\beta_3sin\frac{2\pi x}{12}+\beta_5cos\frac{2\pi x}{\beta_4}+\beta_6sin\frac{2\pi x}{\beta_4}+\beta_8cos\frac{2\pi x}{\beta_7}+\beta_9sin\frac{2\pi x}{\beta_7}$ & 9 & 168\\
8 & MGH09 & $\frac{\beta_1(x^2+x\beta_2)}{x^2+x\beta_3+\beta_4}$ & 4 & 11 \\
9 & Thurber & $\frac{\beta_1+\beta_2x+\beta_3x^2+\beta_4x^3}{1+\beta_5x+\beta_6x^2+\beta_7x^3}$ & 7 & 37 \\
10 & Rat42 & $\frac{\beta_1}{1+exp(\beta_2-\beta_3x)}$ & 3 & 9 \\
\noalign{\smallskip}\hline
\end{tabular}
\end{table}

\subsection{Parameters Setting}
The parameters of MVO are set as the following: Number of universe = $30$, $WEP_{\max} = 1$, $WEP_{\min} = 0.2$, exploitation accuracy ($p$)= $6$, the maximum number of  iterations=$100$.\\
The parameters of PSO are set as the following: population size=$30$, $w_{\max}=0.9$, $w_{\min}=0.4$, $c_1=2.05, c_2=2.05$, 
the maximum number of  iterations=$100$.
\subsection{PC Configuration}

\begin{itemize}
    \item CPU: Intel i3-4005U 1.70GHz
    \item RAM: 4GB
    \item Operating System: Windows 7
    \item Software Tool: MATLAB R2018a
\end{itemize}


\section{Results \& Discussion}
\label{sec:3}
In this work, nonlinear regression analysis has been performed using the MVO algorithm for $10$ regression models. `Hold-out' cross-validation method is used. $80\%$ of the dataset is used in training of the model and the remaining $20\%$ of the  dataset is used as test data for the model. The experiment is repeated $31$ times for each model. The same experiment is conducted using PSO for the comparative study.
The quality of the results has been measured in terms of training and testing MSE errors over 31 independent runs.
The mean and standard deviation of MSE values in training over $31$ runs are given in Table~\ref{tab:training}. The mean and standard deviation of MSE values in testing over $31$ runs are given in Table~\ref{tab:1}.
\begin{table}[!hbt]
\caption{Mean and standard deviation of training residual errors over 31 independent runs.}
\label{tab:training}
\begin{tabular}{lll}
\hline\noalign{\smallskip}
Model & PSO & MVO\\
\noalign{\smallskip}\hline\noalign{\smallskip}
Misra1a & 3.389(0.1696) &\textbf{0.2638(0.1246)}\\
Gauss1 & 80.4833(130.1000) & \textbf{5.5966(0.4977)}\\
Danwood & 0.0076(0.0288) & \textbf{6.05E-04(1.42E-05)}\\
Nelson & 0.0499(0.0441) & \textbf{0.0271(1.37E-04)}\\
Lanczos2 & 5.96E-04(8.37E-04) & \textbf{2.76E-06(5.01E-06)}\\
Roszman1 & 2.62E-05(2.20E-05) & \textbf{1.56E-05(4.42E-07)}\\
ENSO & 11.2482(1.48E-06) & 11.2482(1.82E-05)\\
MGH09 & \textbf{2.58E-05(1.01E-06)} & 2.65E-05(2.86E-07)\\
Thurber & 7.15E+02(8.01E+02) & \textbf{5.46E+02(4.03E+02)}\\
Rat42 & 1.3186(0.4807) & \textbf{0.9483(0.0043)}\\
\noalign{\smallskip}\hline
\end{tabular}
\end{table}

\begin{figure}[!tbh]
\includegraphics[height=6cm, width=13cm]{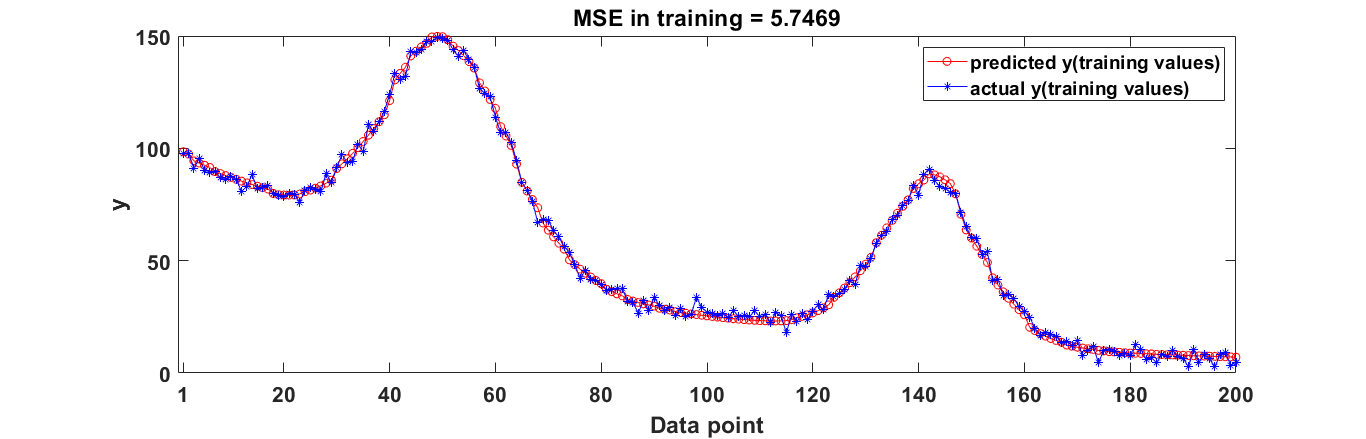}
\caption{Model prediction results using MVO for training data in regression analysis of Gauss1 model.}
\label{fig:1}       
\end{figure}

\begin{figure}[!tbh]
\includegraphics[height=6cm, width=13cm]{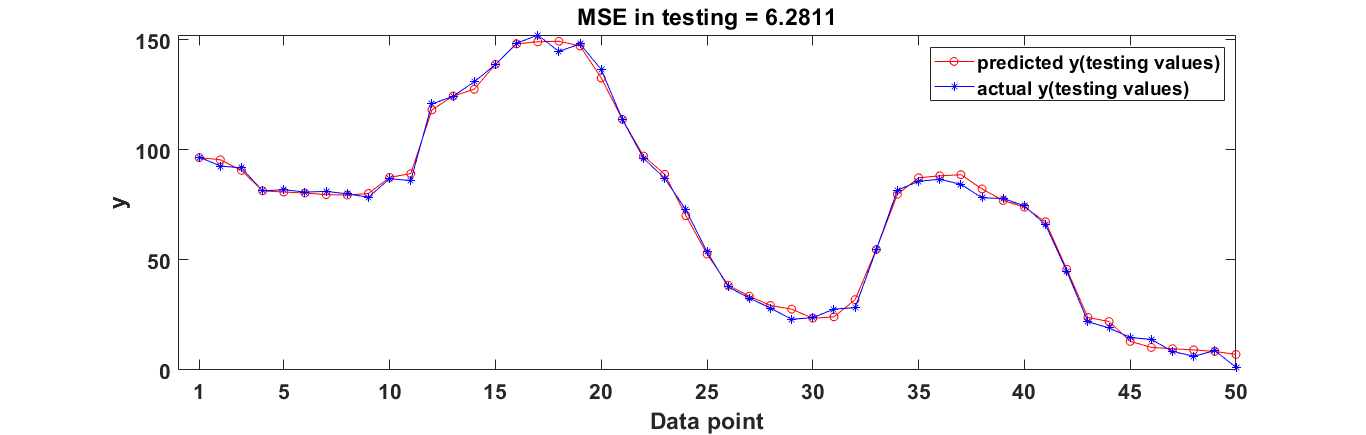}
\caption{Model prediction results using MVO for testing data in regression analysis of Gauss1 model.}
\label{fig:2}       
\end{figure}

\begin{figure}[!tbh]
\includegraphics[height=6cm, width=13cm]{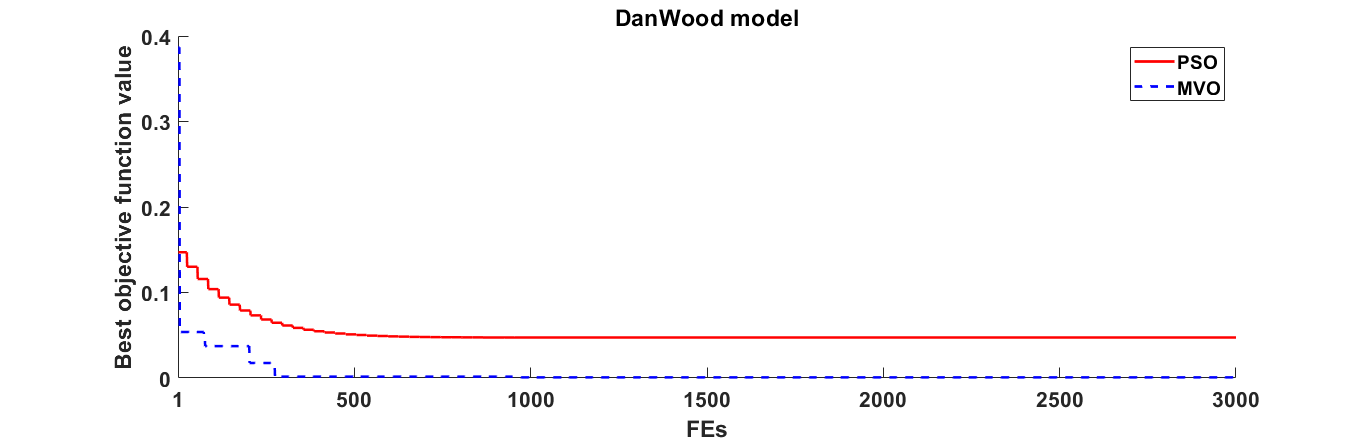}
\caption{Convergence graph of MVO for Danwood model.}
\label{fig:3}       
\end{figure}

\begin{table}[!hbt]
\caption{Mean and standard deviation of testing residual errors over 31 independent runs. $h=1$ indicates MVO statistically outperforms PSO, $h=-1$ indicates PSO statistically outperforms MVO, and $h=0$ indicates no significant difference in the performance.}
\label{tab:1}       
\begin{tabular}{lllll}
\hline\noalign{\smallskip}
Model & PSO & MVO & p-value & h  \\
\noalign{\smallskip}\hline\noalign{\smallskip}
Misra1a & 1.9869(0.1696) & \textbf{0.1381(0.0677)} & 1.17E-06 $\ll$ 0.5 & 1\\
Gauss1 & 76.8515(133.73033) & \textbf{6.2735(0.7319)} & 1.17E-06 $\ll$ 0.5 & 1\\
DanWood & 0.0061(0.0061) & \textbf{0.0013(7.04E-05)} & 5.99E-06 $\ll$ 0.5 & 1\\
Nelson & 0.0707(0.0567) & \textbf{0.0403(5.54E-04)} & 2.12E-04 $\ll$ 0.5 & 1\\
Lanczos2 & 8.39E-04(0.0012) & \textbf{2.41E-06(4.26E-06)} & 2.56E-06 $\ll$ 0.5 & 1\\
Roszman1 & \textbf{2.96E-05(5.61E-06)} & 3.78E-05(2.55E-06) & 3.10E-06 $\ll$ 0.5 & -1\\
ENSO & 13.4705(6.06E-06) & 13.4705(7.45E-05) & 0.9064 $>$ 0.5 & 0\\
MGH09 & 3.91E-05(7.19E-06) & \textbf{3.48E-05(1.18E-06)} & 0.004 $<$ 0.5& 1\\
Thurber & 9.16E+02(1.05E+03) & \textbf{3.57E+02(2.10E+02)} & 3.70E-03 $\ll$ 0.5& 1\\
Rat42 & 2.617(0.3246) & \textbf{0.4734(0.0349)} & 1.17E-06 $\ll$ 0.5& 1\\
 \\

\noalign{\smallskip}\hline
\end{tabular}
\end{table}
From Table~\ref{tab:training}, it is observed that MVO performs better in training than PSO for most of the models. The standard deviations of training MSEs of MVO are also lower than that of PSO. It is observed from Table~\ref{tab:1} that the mean testing MSEs are better than that of PSO for most of the models. To test the significance in the difference of performance of MVO and PSO, a non-parametric statistical test, Wilcoxon's Signed Ranked Test~\cite{Ref11} has been carried out  with significance level ($\alpha$) = $0.05$. The p-values and null hypothesis values ($h$) are given in Table~\ref{tab:1}. The p-values less than $0.05$ indicates  statistically significant difference in the performance  whereas p-values greater than or equal to $0.05$ depict no significant difference in the performance of the algorithms.
In Table~\ref{tab:1}, $h=1$ indicates MVO statistically outperforms PSO, $h=-1$ indicates PSO statistically outperforms MVO, and $h=0$ indicates no significant difference in the performance.
From Table-~\ref{tab:1}, it is observed that the $h$-values come out to be 1 for 8 out of 10 models and it signifies the statistically better performance of MVO over PSO. PSO statistically outperforms MVO for the Roszman1 model. There is no significant difference in the performance of MVO and PSO for ENSO model. 
The robustness of meta-heuristic algorithms is measured in terms of standard deviations. From Table~\ref{tab:training} \&~\ref{tab:1}, it can be observed that the standard deviation of PSO is lower than PSO that indicates that MVO is more robust than PSO in non-linear regression.
Model prediction results using MVO for training  and testing data in regression analysis of the Gauss1 model are given in Fig.~\ref{fig:1} \&~\ref{fig:2} respectively. From these graphs, it is observed that MVO almost fits the training and testing curves for Gauss1 model. The convergence graph of MVO and PSO is given Fig.~\ref{fig:3}. From this graph, it is observed that MVO has better convergence behavior than PSO.

%
%

%

%
\section{Conclusion}
\label{sec:4}
In this paper, MVO is used for nonlinear regression analysis. MVO is applied to search the parameters of different regression models. For the experiment, 10 well-known benchmark regression models are used. A comparative study has been carried out with PSO. The experimental results demonstrate that the proposed  method statistically outperforms PSO in nonlinear regression analysis. In the future, different meta-heuristic algorithms will be studied in nonlinear regression analysis. 


%
%



\end{document}